\documentclass[table,xcdraw]{article} 
\usepackage{iclr2025_conference,times}

\usepackage{amsmath,amsfonts,bm}

\def\eqref#1{equation~\ref{#1}}

\def\1{\bm{1}}

\DeclareMathAlphabet{\mathsfit}{\encodingdefault}{\sfdefault}{m}{sl}
\SetMathAlphabet{\mathsfit}{bold}{\encodingdefault}{\sfdefault}{bx}{n}

\usepackage{hyperref}
\usepackage{url}
\usepackage{amsmath}
\usepackage{amssymb}
\usepackage{amsfonts}
\usepackage{geometry}
\usepackage{multirow} 
\usepackage{booktabs}
\usepackage{graphicx}               
\usepackage{hyperref}               
\usepackage{comment}
\usepackage{subcaption}
\usepackage{graphicx}
\usepackage{wrapfig}
\usepackage[font=small,labelfont=bf]{caption}

\title{\model{}: Boosting Mixture of LoRA Experts Fine-Tuning with a Hybrid Routing Mechanism}

\author{Dengchun Li$^1$, Naizheng Wang$^1$, Zihao Zhang$^1$, Haoyang Yin$^1$, Lei Duan$^1$, Meng Xiao$^2$\\
\textbf{Mingjie Tang}$^1$\\
$^1$School of Computer Science, Sichuan University, Chengdu, China.\\
$^2$Computer Network Information Center, Chinese Academy of Sciences, Beijing, China.\\
\texttt{mikecovlee@163.com, pherenice1125@gmail.com, zzzzh@stu.scu.edu.cn},\\
\texttt{filtee0812@gmail.com, leiduan@scu.edu.cn, shaow@cnic.cn},\\ \texttt{tangrock@gmail.com}}

\newcommand{\model}{\textsc{DynMoLE}}

\newcommand{\tang}[1]
{{\it\small\textcolor{red}{[ {#1}\ --tang ]}}}

\iclrfinalcopy
\begin{document}

\maketitle

\begin{abstract}
Instruction-based fine-tuning of large language models (LLMs) has achieved remarkable success in various natural language processing (NLP) tasks. Parameter-efficient fine-tuning (PEFT) methods, such as Mixture of LoRA Experts (MoLE), combine the efficiency of Low-Rank Adaptation (LoRA) with the versatility of Mixture of Experts (MoE) models, demonstrating significant potential for handling multiple downstream tasks. However, the existing routing mechanisms for MoLE often involve a trade-off between computational efficiency and predictive accuracy, and they fail to fully address the diverse expert selection demands across different transformer layers.
In this work, we propose \model{}, a hybrid routing strategy that dynamically adjusts expert selection based on the Tsallis entropy of the router's probability distribution. This approach mitigates router uncertainty, enhances stability, and promotes more equitable expert participation, leading to faster convergence and improved model performance. Additionally, we introduce an auxiliary loss based on Tsallis entropy to further guide the model toward convergence with reduced uncertainty, thereby improving training stability and performance.
Our extensive experiments on commonsense reasoning benchmarks demonstrate that \model{} achieves substantial performance improvements, outperforming LoRA by 9.6\% and surpassing the state-of-the-art MoLE method, MoLA, by 2.3\%. We also conduct a comprehensive ablation study to evaluate the contributions of \model{}'s key components.
\end{abstract}
    \section{Introduction}
Instruction-based fine-tuning of large language models~\citep{Brown2020LanguageMA, Chowdhery2022PaLMSL, Touvron2023LLaMAOA, Touvron2023Llama2O} for various downstream tasks has achieved remarkable proficiency in natural language processing tasks~\citep{Chung2022ScalingIL, Iyer2022OPTIMLSL, zheng2023judging}. 
To significantly reduce the computational and memory resources required for full parameter fine-tuning, parameter-efficient fine-tuning methods have emerged~\citep{houlsby2019parameter, Li2021PrefixTuningOC, Lester2021ThePO, BenZaken2021BitFitSP, Liu2022FewShotPF}. 
Among these, LoRA~\citep{hu2021lora} has gained popularity due to its ability to reduce substantial computational costs. 
To maintain both cross-task generalization and computational efficiency, a promising solution~\citep{yang2024moral, luo2024moelora, feng2024mixtureofloras} is to design an architecture that combines the resource-efficient features of LoRA with the versatility of Mixture of Experts (MoE) models~\citep{wu2024parameterefficient, dou2024loramoe, gou2024mixture, Liu2023MOELoRAAM, feng2024mixtureofloras}.
These methods are often referred to as Mixture of LoRA Experts (MoLE). The routing mechanisms of these MoLE methods are mostly derived from standard MoE models, where a fixed number of expert networks are activated. 
However, recent studies indicate that the requirements for experts vary across different transformer layers~\citep{gao2024higher, zeng2024adamoe}, suggesting that the routing mechanism requires further modifications to account for these factors.





Current routing mechanisms~\citep{cai2024survey} can be broadly classified into two categories: 
\textit{1) Soft Routing}: These methods activate all expert networks for each input token, which typically leads to improvement in prediction accuracy~\citep{ma2018modeling,nie2021evomoe,wu2024mixture,dou2024loramoe,pan2024dense}. 
However, this comes at the cost of significant computational overhead, as all experts are involved in the computation~\citep{shazeer2017outrageously}. 
\textit{2) Sparse Routing}: These approaches enhance model efficiency by activating only a subset of experts~\citep{shazeer2017outrageously}. 
Some techniques route each token to a single expert~\citep{fedus2022switch}, while others activate multiple experts, such as using Top-K~\citep{zhou2022mixture} or Top-P~\citep{huang2024harder}, or employing uncertainty-based routing~\citep{wu2024gw}. 
Although sparse routing improves parameter efficiency, it often results in an imbalanced workload among experts, making it necessary to include an auxiliary loss functions to ensure the balance.
Though both of these routing techniques aim to select the optimal set of experts for each input token, neither provides a fully comprehensive solution that accounts for the diverse and complex factors affecting model performance, which raises a key question: \textit{How can we design a \textbf{hybrid routing approach} that considers these factors holistically to provide a more complete solution for MoE and MoLE models?}

\begin{figure}[t]
\centering
\includegraphics[width=0.75\linewidth]{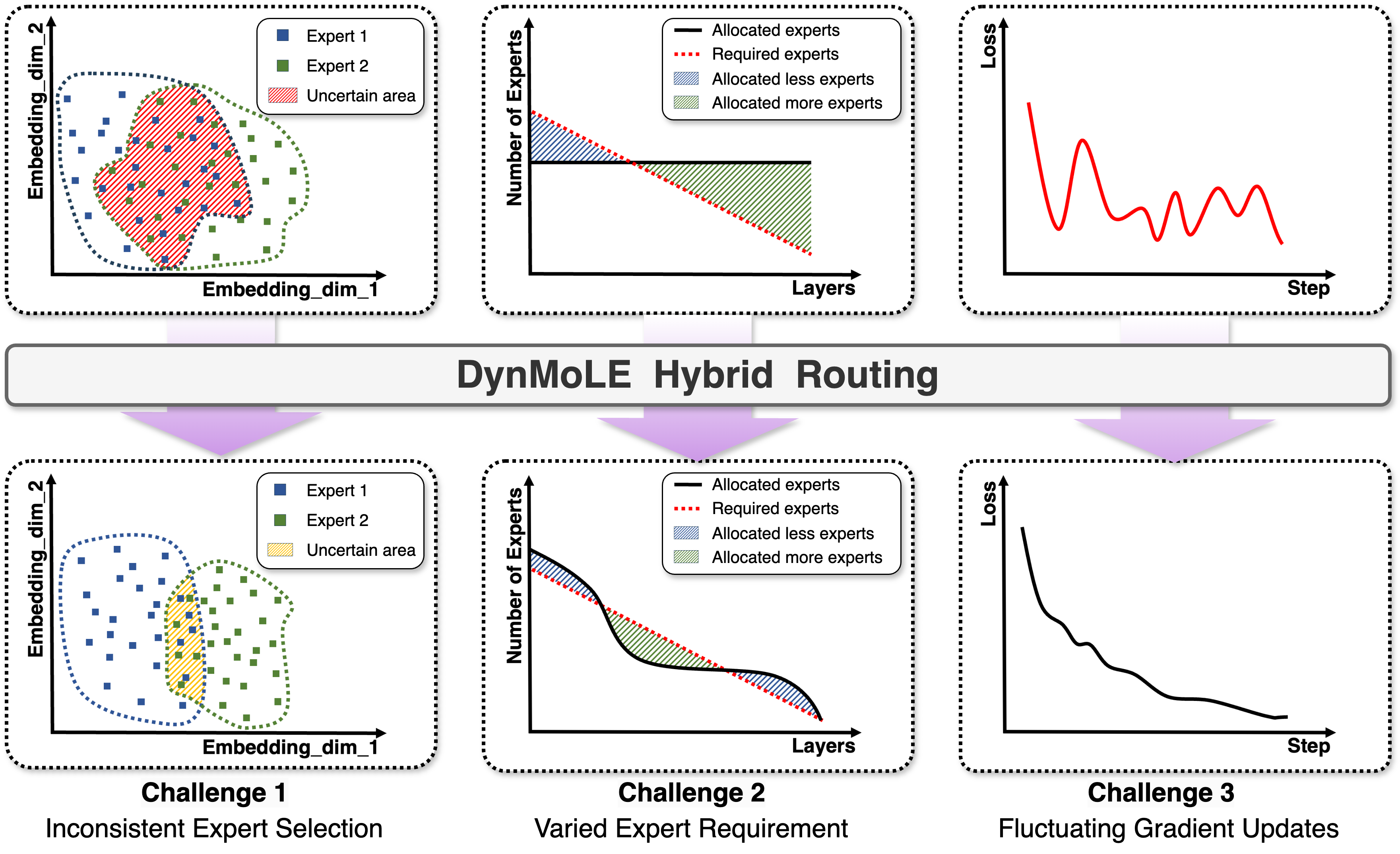}  
\caption{Visualized motivation of \model{}. We propose a \textbf{hybrid routing} mechanism for \model{} to address and solve these critical challenges.}
\label{fig:motivation}
\end{figure} 

To answer this question, several critical challenges emerge: 
\textbf{1) Inconsistent Expert Selection} occurs when flat probability distributions lead to similar inputs activating different experts, resulting in unstable expert training. 
\textbf{2) Varied Expert Requirements} across the model, as noted by~\cite{gao2024higher, zeng2024adamoe}, leads to uneven expert loads when the number of activated experts is fixed across all layers. 
Finally, \textbf{3) Fluctuating Gradient Updates} resulting from uncertain routing decisions, cause fluctuations in gradient flows, which adversely affect convergence speed and stability during model training.
These challenges are illustrated in Figure~\ref{fig:motivation}.

To address these challenges, we propose \model{} (\textbf{Dyn}amic Routing for \textbf{M}ixture \textbf{o}f \textbf{L}oRA \textbf{E}xperts), a hybrid routing approach designed to reduce router uncertainty in Mixture of LoRA Experts adapters for parameter-efficient fine-tuning of large language models. 
Our approach leverages the mathematical properties of Tsallis entropy~\citep{tsallis1988possible}, a generalized entropy measure, to develop adaptive routing strategies that effectively minimize router uncertainty. 
Furthermore, we introduce an auxiliary loss based on Tsallis entropy to guide the model towards convergence with reduced uncertainty, thus improving training stability and performance.
By preventing over-reliance on certain experts and promoting more equitable engagement across all experts, this method fosters a diverse and robust set of expert contributions. This approach not only optimizes computational resource allocation but also enhances overall model performance by improving decision consistency and stability during training.


\textbf{Summary of Contributions}:
\begin{enumerate}
    \item We identify the uncertainty problem in MoE routers and theoretically derive their optimal probability distribution, which we term the \textbf{Peaked Distribution}. Through formal reasoning, we prove that Tsallis entropy provides a more effective quantification of routing uncertainty compared to traditional measures.
    \item We propose a hybrid strategy, called \model{}, which enables the routing mechanism to dynamically adjust based on the entropy of the routing distribution for each token, making expert selection more flexible and efficient. Additionally, we introduce an auxiliary loss for \model{} based on T sallis entropy, to guide the model toward convergence with reduced uncertainty, improving training stability and performance.
    \item We validate the effectiveness of \model{} using widely recognized benchmarks, as used in prior works. The results demonstrate that \model{} achieves remarkable performance, outperforming LoRA by 9.6\% and MoLA, a state-of-the-art MoLE method, by 2.3\%. Furthermore, we conducted a comprehensive ablation study to explore the effectiveness of \model{}'s key components.
\end{enumerate}

\section{Background}

In this section, we introduce the background of the Mixture of Experts (MoE) and Mixture of Large Experts (MoLE), review existing popular routing strategies, and provide the mathematical definitions of uncertainty and entropy.

\subsection{Mixture-of-Experts}

First introduced in 1991 by \citet{jacobs1991moe}, the Mixture-of-Experts (MoE) architecture has seen a resurgence in the context of modern large-scale models, largely attributed to the work of \citet{shazeer2017outrageously}, \citet{mustafa2022multimodal}, \citet{lepikhingshard}, and \citet{fedus2022switch}. Their contributions have made MoE a promising approach for scaling models without significantly increasing computational overhead.

Each MoE layer consists of \( N \) independent networks, referred to as experts, denoted by \( \{E_i\}_{i=1}^{N} \), along with a gating function \( G \), which assigns weights to each expert based on a probability distribution. The forward propagation process \( f_{\text{MoE}} \) for a given input \( x \) can be mathematically expressed as:

\begin{equation}
    f_{\text{MoE}}(x) = \sum_{i=1}^{N} G(x)_i \cdot E_i(x),
\label{eq:vanilla_moe}
\end{equation}

where router logits \( G(x)_i \) represent the routing probabilities for each expert. Each expert \( E_i(x) \) produces an output based on the input \( x \), allowing the MoE architecture to leverage the strengths of multiple specialized models.

In this paper, we focus exclusively on routing strategies for MoLE due to the computational resource limitations of our team. The pre-training and fine-tuning of MoE models require computational resources beyond our current capacity.

\subsection{Mixture-of-LoRA-Experts}

The MoLE architecture \citep{wu2024mixture} extends the traditional Mixture of Experts approach by integrating Low-Rank Adaptation into expert layers, significantly enhancing computational efficiency. In a MoLE layer, each expert $E_i(\cdot)$ is a LoRA-enhanced module that updates only a subset of parameters while leveraging the pretrained knowledge of the base model.

Low-Rank Adaptation, introduced by \citet{hu2021lora}, proposes a method that adjusts only a small number of additional parameters, rather than updating the entire weight matrix of the model. A LoRA block consists of two matrices, $B \in \mathbb{R}^{d \times r}$ and $A \in \mathbb{R}^{r \times k}$, where $d$ and $k$ represent the dimensions of the pretrained weight matrix $W_0 \in \mathbb{R}^{d \times k}$ in large language models. The parameter $r$ is the low-rank dimension, with $r \ll \min(d, k)$. The updated weights $W'$ are computed as:

\begin{equation}
    W' =  W_0 + \Delta W = W_0 + BA,
\label{eq:lora}
\end{equation}

where $\Delta W = BA$ represents the LoRA-induced weight update. Formally, given $N$ experts in a MoLE layer, denoted by $\{E_i\}_{i=1}^{N}$, and router logits $G(x)_i$ representing the routing probabilities for each expert, the forward propagation process $f_{\text{MoLE}}$ for a given hidden state $x$ is calculated as:

\begin{equation}
    f_{\text{MoLE}}(x) = W_0x + \Delta W x = W_0x + \sum_{i=1}^{N} G(x)_i \cdot E_i(x),
\label{eq:mole}
\end{equation}

where $W_0$ denotes the pretrained weights of the base model, and $\Delta W$ represents the weight updates generated by the LoRA-enhanced experts. Each expert $E_i(x)$ computes its output using the LoRA update rule:

\begin{equation}
    E_i(x) = B_i A_i x,
\end{equation}

with $B_i \in \mathbb{R}^{d \times r}$ and $A_i \in \mathbb{R}^{r \times k}$. The low-rank matrix multiplication significantly reduces the number of trainable parameters, improving memory efficiency and accelerating fine-tuning compared to standard MoE architectures. By incorporating the parameter-efficient updates of LoRA, MoLE significantly enhances both computational efficiency and model performance, especially in scenarios that require fine-tuning across multiple tasks.

\subsection{Routing Algorithms}\label{RA}

\paragraph{Softmax Routing} As a classic non-sparse gating function~\citep{jordan1994hierarchical}, it involves multiplying the input by a trainable weight matrix $W_g$ and then applying the $\text{softmax}$ function.

\begin{equation}
    G(x) = \text{softmax}(x \cdot W_g)
\end{equation}

This standard routing mechanism, often referred to as \textbf{soft routing}, is the foundation of all MoE routing algorithms, enabling the MoE model to adaptively allocate resources based on the specific requirements of the input.

\paragraph{Top-K Routing} This is one of the most commonly used MoE routing algorithms. Let $\mathbf{R}(x) = \text{sort}(G(x))$ represent the sorted probability distribution over experts for the input token $x$, and the algorithm is then defined as follows:

\begin{equation}
    \text{Top}_k(\mathbf{R}(x)) = \{ i \mid p_i(x) \geqslant p_{(k)}(x) \},
\label{eq:topk}
\end{equation}

where $p_i(x)$ is the probability assigned to expert $i$, and $p_{(k)}(x)$ is the $k$-th highest probability in $\mathbf{R}(x)$. Top-k routing selects the $k$ experts with the highest probabilities, balancing efficiency and performance.


\paragraph{Top-P Routing} Introduced by \citet{huang2024harder}, this algorithm aims to achieve variability in expert selection. It activates experts dynamically by selecting those whose cumulative probability exceeds the given threshold. This flexibility allows for tailored expert activation based on each token. The Top-p algorithm is formulated as follows:

\begin{equation}
    \text{Top}_p(\mathbf{R}(x)) = \{ i \mid \sum_{j=1}^{i} p_{(j)}(x) \geqslant p \},
\label{eq:topp}
\end{equation}

where $p_{(j)}(x)$ denotes the $j$-th highest probability in the sorted probability distribution $\mathbf{R}(x)$, and $p$ is the cumulative probability threshold. The algorithm selects the smallest set of experts whose cumulative probability is at least $p$, enhancing both efficiency and adaptability.

\subsection{Uncertainty and Entropy} 




Entropy was first introduced by~\citet{shannon1948mathematical} to quantify the amount of "choice" involved in the selection of an event, or the level of uncertainty in its probability distribution. The Shannon entropy is defined as:

\begin{equation}
    H(p) = -\sum_{i=1}^{N} p_i \log p_i,
    \label{eq:shannon_entropy_def}
\end{equation}

where \( p = \{p_1, p_2, \ldots, p_N\} \) represents a probability distribution over \( N \) events.

Shannon entropy has been widely applied in various fields, such as natural language processing and machine learning~\citep{jelinek1980interpolated, quinlan1986induction}. However, \citet{alomani2023further} argues that the non-additive property of Tsallis entropy~\citep{tsallis1988possible} provides an advantage over Shannon entropy in handling complex systems and non-Gaussian distributions. Tsallis entropy introduces a tunable parameter \( q \), which offers greater flexibility in measuring uncertainty under different conditions. The parameter \( q \), known as the entropic index, controls the degree of non-extensivity. The Tsallis entropy is defined as:

\begin{equation}
    S_q(p) = \frac{1}{q - 1} \left( 1 - \sum_{i=1}^{N} p_i^q \right),
    \label{eq:tsallis_entropy_def}
\end{equation}








\section{Deep Dive into the Routing Mechanism}

In this section, we present an in-depth analysis of the routing mechanism employed in MoE and MoLE architectures.

\subsection{What is the ideal distribution of routing weights?}

Given an $N$-expert MoE model using a soft routing algorithm, the router generates a probability distribution normalized by Softmax function. The distribution corresponds to the proportion of outputs from each expert. The initial distribution \( G(x) \) is set as uniform. It is optimized during training to minimize a loss function \( L(f_{\text{MoE}}(x), y) \) which is convex and differentiable. \( f_{\text{MoE}}(x) \) is the model's predicted output for input \( x \), and \( y \) is the true label. From \ref{appd:muomr}, we proof that the gradient of the loss function \(L\) with respect to $G_i(x)$ is proportional to the expert output $E_i(x)$, that is:

\begin{align}
\frac{\partial L}{\partial G_i(x)} \propto E_i(x)
\end{align}

This indicates that the variation of \( G_i(x) \) during training is highly dependent on the contributions of the experts to the model’s output, where the expert outputs directly influence the direction and magnitude of the weight updates. In other words, for experts who contribute to a reduction in loss, the gating network increases their weights; For other experts who result in an increase in loss, it decreases their weight.

Ideally, when the model completely converges, the weights that $G(x)$ assigned to the subset of experts who contribute most significantly to loss reduction, will be close to 1. This causes the distribution of \( G(x) \) to approach an \textbf{indicative~distribution}, expressed in a characteristic function form as follows:

\begin{align}
G_i(x) = \begin{cases} 
1, & i \in \underset{j}{\arg\min} \, L(E_j(x), y)\\ 
0, & \text{otherwise} 
\end{cases}
\end{align}

Though in practice, due to regularization terms and numerical stability, $G(x)$ cannot form a perfect indicative distribution, the overall trend of allocating higher weights to a smaller subset of experts, will form a \textbf{peaked distribution}, as shown in Figure~\ref{fig:main_pic}. We use \textbf{uncertainty} to describe how much the current distribution deviates from the ideal peaked distribution (indicative distribution). A greater uncertainty indicates a more uniform distribution and higher confusion in the router selection. However, since it is hard to define peaked distribution directly in an analytic expression mathematically, traditional measures such as KL divergence based on mutual information, are inadequate. Therefore, we introduce the concept of \textbf{entropy} from information theory.

\subsection{Tsallis Entropy vs. Shannon Entropy}

\paragraph{Tsallis Entropy Provide More Flexible.}
As \(q\to1\), the Tsallis entropy degenerates to the Shannon entropy:
\begin{align}
\lim_{q\to 1}S_q(p) = \lim_{q\to 1} \frac{1 - \sum_{i=1}^N p_i^q}{q - 1} = -\sum_{i=1}^N p_i \log p_i = H(p)
\end{align}

The Tsallis entropy provides a continuous framework that includes the Shannon entropy as a special case. The entropic-index $q$ acts as a tunable hyperparameter, offering a powerful tool. By adjusting $q$, we can modulate the sensitivity of the entropy to the probability distribution, making it a more adaptable and flexible tool across various scenarios. The detail information is at \ref{appd:flex}



\paragraph{Tsallis Entropy Provide More Training Stability.}
Consider a loss function that incorporates an entropy regularization term: $\mathcal{L} = \mathcal{L}_{\text{data}} - \lambda \cdot \text{Entropy}(f_{\text{MoE}}(x))$, where $\lambda$ is the regularization coefficient and $\text{Entropy}(f_{\text{MoE}}(x))$ quantifies the uncertainty in the router's output. Now, let us compare the loss functions using Shannon entropy and Tsallis entropy, respectively. Let $G_i(x)$ represent the routing probability of expert $i$:

\begin{align}
    \mathcal{L}_{\text{Shannon}} &= \mathcal{L}_{\text{data}} - \lambda \cdot \sum_{i=1}^n G_i(x) \log G_i(x) \\
    \mathcal{L}_{\text{Tsallis}} &= \mathcal{L}_{\text{data}} - \lambda \cdot \frac{1}{q - 1} \left( 1 - \sum_{i=1}^n G_i(x)^q \right)
\end{align}

\begin{wrapfigure}[18]{}{0.28\textwidth}
    \centering
    \includegraphics[width=\linewidth]{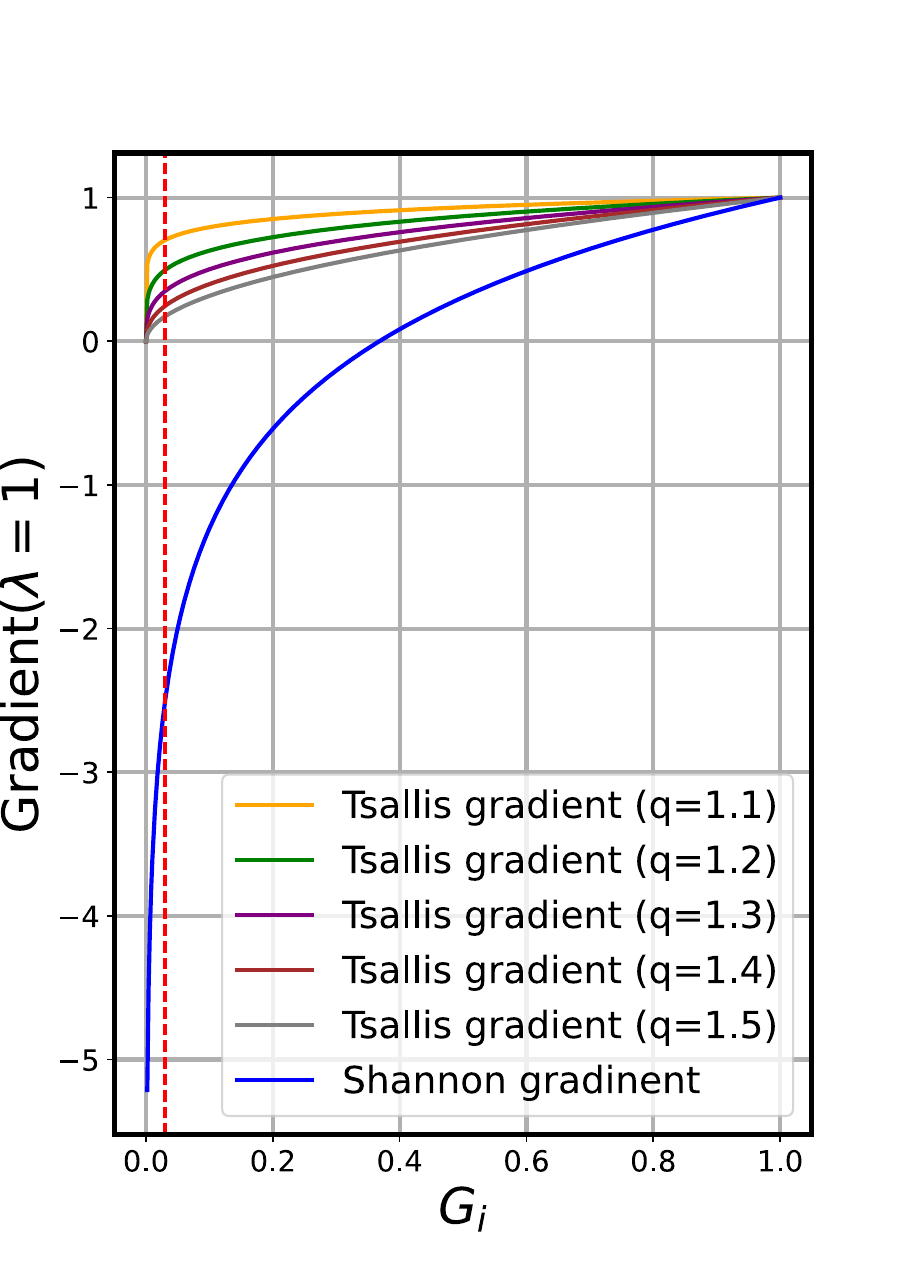}
    \caption{Tsallis entropy provides a more stable optimization process than Shannon entropy by reducing the impact of low-probability events.}
    \label{fig:grad}
\end{wrapfigure}

When optimizing these loss functions via gradient descent, the gradients with respect to $G_i(x)$ are given by:

\begin{align}
    \frac{\partial \mathcal{L}_{\text{Shannon}}}{\partial G_i} =& \frac{\partial \mathcal{L}_{\text{data}}}{\partial G_i} - \lambda \cdot\left(1 + \log G_i\right) \\
    \frac{\partial \mathcal{L}_{\text{Tsallis}}}{\partial G_i} =& \frac{\partial \mathcal{L}_{\text{data}}}{\partial G_i} - \lambda \cdot G_i^{q - 1}
\end{align}

Figure~\ref{fig:grad} clearly shows the trend of gradients of the two entropy functions as $G_i(x)$ changes. For Shannon entropy, as $G_i(x) \to 0$, $\log G_i \to -\infty$, which can lead to steep gradient magnitudes and unstable updates. In contrast, for Tsallis entropy, as $G_i(x) \to 0$, the gradient $\lambda G_i^{q - 1} \to 0$ ($q > 1$), reducing the impact of low-probability events and providing a more stable optimization process.

\textbf{Tsallis Entropy Provide More Certainty:}
As $q$ increases, Tsallis entropy assigns greater weights to high-probability events in the entropy calculation, while the contribution from low-probability events diminishes. This encourages the optimization process to choose a few experts with higher probabilities. As a result, the model is biased towards reliable experts and avoids the uncertain ones, effectively reducing the overall uncertainty in the decision-making process.

\clearpage

\section{\model{}}

\begin{figure}[t]
\begin{center}
\includegraphics[width=0.85\linewidth]{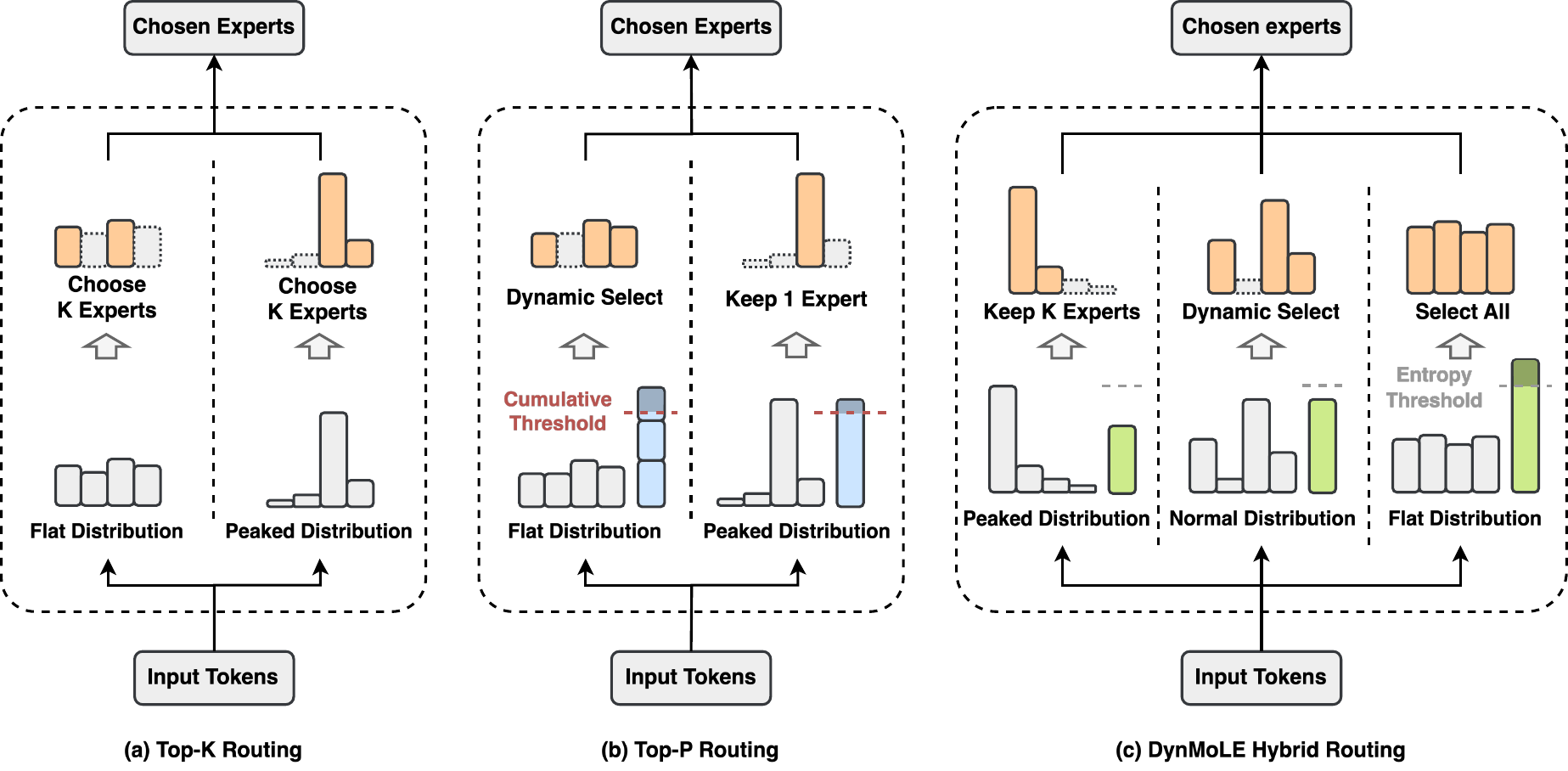}  
\end{center}
\caption{Comparison of three routing strategies: (a) the classic Top-K Routing, here we use Top-2 as example; (b) the classic Top-P Routing, where the blue bars represent the sum of the highest probabilities; and (c) \model{} Hybrid Routing, where the green bars represent the entropy values across different probability distributions. 
}
\label{fig:main_pic}
\end{figure}

In this section, we present the routing algorithm used in \model{}, which integrates entropy-based selection across various routing strategies to efficiently allocate tokens to experts, as shown in Figure~\ref{fig:main_pic}(c). This approach, termed \textbf{dynamic routing}, leverages Tsallis entropy to dynamically switch between soft routing, top-$p$ routing, and top-$k$ routing mechanisms. By assigning the most suitable experts to each token, we reduce routing uncertainty and improve model efficiency. Additionally, we incorporate Tsallis entropy into an auxiliary loss to guide the model towards convergence with reduced uncertainty.

\subsection{Entropy-based Intelligent Hybrid Routing}

In this study, we use Tsallis entropy to capture the deviation from the ideal peaked distribution, providing greater flexibility than KL divergence. While Top-k and Top-p routing are effective, both struggle when the router has high uncertainty, leading to nearly uniform probability distributions and reducing expert prioritization, which results in suboptimal performance.

To address this, we propose \model{}, a hybrid strategy that dynamically adjusts routing based on the entropy of each token. For high-entropy tokens, indicating greater uncertainty, we use \textbf{soft routing}, allowing the model to select from a broader set of experts. For low-entropy tokens, a more deterministic \textbf{Top-p} routing is applied to focus on a narrower set of likely experts. Additionally, at least $k$ experts are always activated to prevent overfitting. This dynamic approach balances exploration and exploitation, improving performance across different conditions.

Given the sorted router probabilities for the input token $x$ as $\mathbf{R}(x) = \text{sort}(G(x))$, and the Tsallis entropy of the router probability as $\mathbf{S}(x) = S_q(\mathbf{R}(x))$, with a routing threshold $H_{\text{threshold}}$, the hybrid routing $G_{\text{hybrid}}(x)$ can be expressed as:

\begin{align}
    G_{\text{hybrid}}(x) =
\begin{cases}
    \mathbf{R}(x), & \text{if } \mathbf{S}(x) > H_{\text{threshold}} \\
    \text{Top}_{(p,k)}(\mathbf{R}(x)), & \text{otherwise}
\end{cases}
\end{align}

where $\text{Top}_{(p,k)}(\mathbf{R}(x))$ is defined as:

\begin{align}
    \text{Top}_{(p,k)}(\mathbf{r}(x)) =
\begin{cases}
    \text{Top}_p(\mathbf{R}(x)), & \text{if } p_{(j)}(x) \geqslant p_{(k)}(x) \\
    \text{Top}_k(\mathbf{R}(x)), & \text{otherwise}
\end{cases}
\end{align}
where $p_{(j)}(x)$ represents the $j$-th largest probability in $\mathbf{r}(x)$ for the Top-p selection, and $p_{(k)}(x)$ denotes the $k$-th largest probability for the Top-k selection. The function $\text{Top}_p(\mathbf{r}(x))$ selects the Top-p fraction of probabilities, while $\text{Top}_k(\mathbf{r}(x))$ ensures that at least $k$ experts are selected, guaranteeing sufficient diversity in expert participation based on router probabilities.

\subsection{Auxiliary Entropy Loss}\label{AEL}

To further reduce the uncertainty of the router and promote balanced expert usage, we introduce an auxiliary loss based on Tsallis entropy and load balancing. Given $N$ experts, the sorted router probabilities for the input token $x$ are denoted by $\mathbf{R}(x) = \text{sort}(\text{softmax}(G(x)))$, with the Tsallis entropy of the router probability defined as $\mathbf{S}(x) = S_q(\mathbf{R}(x))$. For a batch $\mathcal{B}$ containing $T$ tokens, the entropy loss is computed as:
\begin{align}\label{eq:entropy_loss}
    L_{\text{entropy}} = \beta \cdot \frac{1}{T} \sum_{x \in \mathcal{B}} \mathbf{S}(x),
\end{align}
where $\beta$ is a multiplicative coefficient controlling the impact of the entropy loss.

To encourage a balanced load across experts, we also introduce a load balance loss from~\cite{fedus2022switch} defined as:
\begin{align}
    L_{\text{balance}} = \alpha \cdot N \cdot \sum_{i=1}^{N} f_i \cdot P_i
\end{align}




The overall auxiliary loss combines both the entropy loss and the load balance loss:

\begin{align}\label{eq:aux_loss}
    L_{\text{auxiliary}} = L_{\text{balance}} + L_{\text{entropy}}
\end{align}

By incorporating this auxiliary loss, \model{} improves model performance by addressing both router uncertainty (through the Tsallis entropy loss) and router imbalance (through the load balance loss), leading to more efficient expert utilization and reduced uncertainty in routing decisions.

\clearpage

\section{Experiments}

\begin{table}[!t]
    \centering
    \caption{Performance comparison of LLaMA-2-7B models with different PEFT methods across various benchmarks.}
    \renewcommand{\arraystretch}{1.5}
    \scalebox{0.84}{ 
    \begin{tabular}{lcccccccccc}
        \toprule
        \textbf{PEFT Method} & \textbf{RTE} & \textbf{ARC-e} & \textbf{ARC-c} & \textbf{BoolQ} & \textbf{OBQA} & \textbf{PIQA} & \textbf{SIQA} & \textbf{HellaS} & \textbf{WinoG} & \textbf{AVG.} \\ 
        \midrule
        LoRA & 52.7 & 73.8 & 50.9 & 68.2 & 77.4 & 81.1 & 69.9 & 88.4 & 68.8 & 70.1 \\ 
        DoRA & 52.7 & 76.5 & 52.8 & 71.7 & 78.6 & \textbf{82.7} & 74.1 & 89.6 & 69.3 & 72.0 \\ 
        LoRAMoE (Soft Routing) & 55.6 & 75.7 & 51.5 & 71.7 & 78.4 & 81.9 & 77.7 & 93.5 & 75.6 & 73.5 \\ 
        MoLA (Top-K) & 69.3 & 76.7 & 52.4 & 72.3 & 78.2 & 82.0 & \textbf{78.7} & 93.2 & 75.1 & 75.3 \\ 
        \model{} (Top-P) & 70.8 & 76.1 & 52.6 & 72.9 & 76.2 & 82.1 & 78.1 & \textbf{93.7} & \textbf{77.8} & 75.6 \\ 
        \midrule
        \textbf{\model{}(-)} & 69.0 & 76.8 & 54.9 & \textbf{73.5} & 78.0 & 81.9 & 78.1 & 93.2 & 76.7 & \textbf{75.8} \\ 
        \textbf{\model{}} & \textbf{80.1} & \textbf{78.6} & \textbf{56.0} & 72.9 & \textbf{79.2} & 82.5 & 78.5 & 92.9 & \textbf{77.8} & \textbf{77.6} \\ 
        \bottomrule
    \end{tabular}}
    \label{table:main_result}
    \begin{flushleft}
        \hspace{0.8cm}
        (-) Indicates that \model{} trained without auxiliary entropy loss.
    \end{flushleft}
\end{table}

In this section, we present a comprehensive evaluation of \model{}. We compare the performance of \model{} with other state-of-the-art fine-tuning methods. Our experiments are designed to assess the generalization capability of \model{} in handling diverse tasks. Through extensive comparisons, we demonstrate that \model{} consistently outperforms baseline methods in terms of accuracy, particularly when integrated with entropy-based routing. Additionally, we show that \model{} achieves superior performance while maintaining parameter efficiency, highlighting its effectiveness in large language model fine-tuning.

\subsection{Experimental Setup}

\noindent\textbf{Datasets.} To evaluate the effectiveness of \model{}, we conducted experiments on a diverse set of commonsense reasoning datasets, following prior work~\citep{liu2024dora, li2024mixlora}. The datasets are as follows: ARC\citep{clark2018think}, OpenBookQA\citep{mihaylov2018obqa}, PIQA\citep{bisk2020piqa}, SocialIQA\citep{sap2019socialiqa}, BoolQ\citep{clark2019boolq}, Hellaswag\citep{zellers2019hellaswag}, Winogrande\citep{sakaguchi2021winogrande}, and GLUE\citep{wang2018glue}. These datasets provide a comprehensive assessment of LLMs across various challenges, ranging from scientific queries to commonsense inference. The performance of all methods are measured using accuracy across all datasets. Further details are provided in Appendix~\ref{appd:dataset}.

\noindent\textbf{Baselines.} In line with previous studies~\citep{dou2024loramoe, gao2024higher}, we employed the widely-adopted Llama-2-7B as the base model. To thoroughly assess the performance of \model{}, we compared it against several prominent parameter-efficient fine-tuning (PEFT) methods, 
including LoRA\citep{hu2021lora}, DoRA\citep{liu2024dora}, LoRAMoE\citep{dou2024loramoe} (representing \textbf{soft routing}), and MoLA\citep{gao2024higher} (representing \textbf{Top-K routing}).
While no existing PEFT methods explicitly use the \textbf{Top-P routing} strategy\citep{huang2024harder}, we fixed \model{}'s routing algorithm to Top-P to evaluate the performance of all fundamental routing strategies.

\noindent\textbf{Settings.} To ensure parameter consistency across experiments, both LoRA and DoRA are initialized with a rank of $r=80$, while LoRAMoE and MoLA are initialized with a rank of $r=16$ across 6 experts. For all baselines, we apply updates to the \(gate\_proj\), $down\_proj$, and $up\_proj$ weights within the feed-forward network (FFN) layers to ensure fair comparisons. Importantly, we control the number of trainable parameters across all methods, ensuring that \model{} and other MoE-based approaches have an identical number of trainable parameters—approximately \textbf{3\% (200 million)} of the total model parameters. Further details on hyperparameters are provided in Appendix~\ref{appd:hyper-p}.

\subsection{Main Results}
Table \ref{table:main_result} provides a comprehensive comparison of various PEFT methods, applied to the LLaMA-2-7B model across a diverse set of benchmarks. \model{} consistently demonstrates superior performance, especially when combined with entropy loss, achieving an average accuracy of 77.6\%, which surpasses all baseline methods. These results underscore the effectiveness of incorporating Tsallis entropy as a measure to improve routing decisions in MoE-based architectures.
For individual tasks, \model{} with entropy loss exhibits remarkable improvements in challenging benchmarks such as ARC-c and PIQA, achieving 56.0\% and 82.5\%, respectively. In particular, \model{} outperforms traditional PEFT method LoRA by 7.5\% and state-of-art PEFT method DoRA by 4.6\%, clearly demonstrating the superiority of the MoLE.

To further validate the efficacy of our proposed hybrid routing strategy, We included three important baselines: LoRAMoLE, MoLA, and \model{} (Top-p), representing soft routing, Top-k routing, and Top-p routing, respectively. The Top-p method is effectively implemented by disabling the soft routing mechanism, eliminating the entropy loss calculation of \model{}, and reducing the minimum number of activated experts(Refers to the super parameter Keep-Top-k) to one. LoRAMoE, as a soft MoE method, achieved an average accuracy of 73.5\%. It combines the strengths of the LoRA module and the MoE architecture, leading to considerable improvements over traditional PEFT methods, yet there is still significant room for improvement in enhancing the specialization of its experts. While MoLA achieves an average accuracy of 75.3\%, showcasing that although Top-k routing is competitive, it struggles to dynamically adjust the number of active experts based on token uncertainty. On the other hand, \model{} (Top-p) delivers a commendable performance with an average accuracy of 75.6\%, but its pure Top-p routing mechanism does not fully exploit the flexibility required for dynamic token-expert assignment (Detailed results are shown in Appendix~\ref{appd:flex_studies}).

The comparisons above highlight the advantage of \model{}'s hybrid routing strategy, which leverages the benefits of both Top-p and Top-k mechanisms. \model{}(-) leverages Tsallis entropy to assist the router in customizing token routing, surpassing other advanced MoLE methods by more than 0.2\%. Notably, by integrating a newly designed auxiliary entropy loss, \model{} optimizes both of the router uncertainty and load balancing among experts more effectively, maintaining an accuracy advantage of over 2\% compared to other MoLE methods. In particular, in benchmarks like ARC-c and OBQA, the performance gap is narrower, yet \model{} still maintains a consistent lead. This consistent performance across tasks highlights \model{}'s strong generalization ability, even in scenarios where the distinction between models is less pronounced.


\subsection{Ablation Studies}

\begin{figure}[!t]
    \centering
    \begin{minipage}{0.025\linewidth}
        \includegraphics[width=\linewidth]{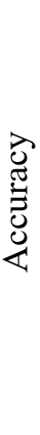}
    \end{minipage}%
    \begin{minipage}{0.18\linewidth}
        \includegraphics[width=\linewidth]{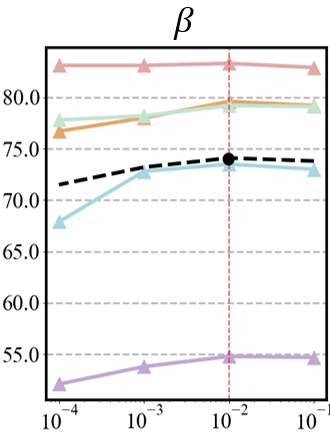}
        \subcaption{}
    \end{minipage}%
    \begin{minipage}{0.18\linewidth}
        \includegraphics[width=\linewidth]{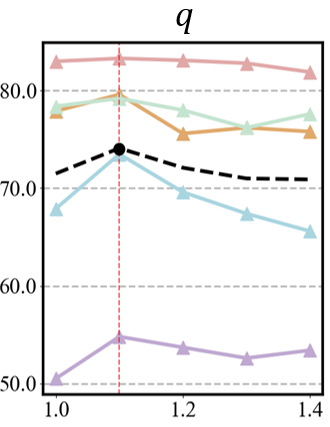}
        \subcaption{}
    \end{minipage}%
    \begin{minipage}{0.18\linewidth}
        \includegraphics[width=\linewidth]{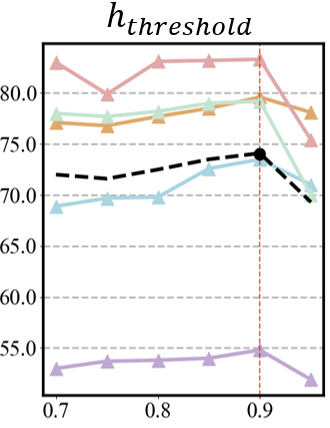}
        \subcaption{}
    \end{minipage}%
    \begin{minipage}{0.18\linewidth}
        \includegraphics[width=\linewidth]{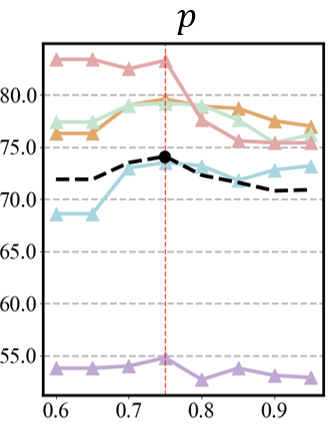} 
        \subcaption{}
    \end{minipage}%
    \begin{minipage}{0.18\linewidth}
        \includegraphics[width=\linewidth]{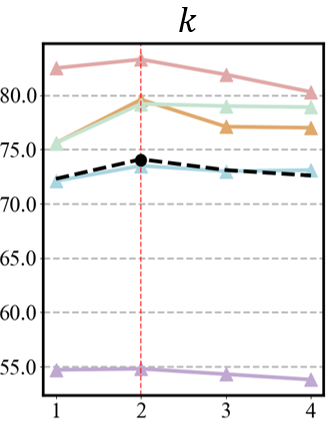}
        \subcaption{}
    \end{minipage}
    \begin{minipage}{0.040\linewidth}
        \includegraphics[width=\linewidth]{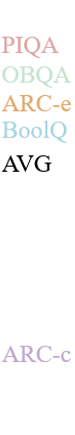}
    \end{minipage}
    \caption{The result of ablation studies: (a) analyzes the effect of varying the proportion of Entropy Router Loss, (b) describes the impact of the Tsallis Entropic Index $q$ on model performance (when $q=1$, it becomes Shannon entropy), (c) reveals the results of different soft routing thresholds, (d) shows the impact of the Top-p threshold, and (e) presents the analysis of Keep-Top-k.}
    \label{fig:images}
\end{figure}

\noindent In this section, we present a comprehensive ablation study to analyze the impact of various key hyperparameters on the performance of \model{}, across the ARC, OpenBookQA, BoolQ, and PIQA datasets using LLaMA-2-7B. The results of these ablations, summarized in Figure~\ref{fig:images}.

\subsubsection{Impact of Different Entropy Settings on \model{}'s Performance}

In this part, we primarily discuss the impact of three entropy-related factors on model performance: 

\textbf{Entropy Loss Coefficient} refers to a key parameter that balances the proportions of Tsallis entropy loss and load balance loss (Equation \ref{eq:entropy_loss}). We designed different \(\beta\) values ranging from \(1 \times 10^{-4}\) to \(1 \times 10^{-1}\). Figure \ref{fig:images}(a) shows that our findings indicated an entropy router loss coefficient of \(1 \times 10^{-2}\) achieved the highest average accuracy, effectively addressing the challenges of imbalanced expert selection. Conversely, disabling the entropy router loss or employing excessively high coefficients disrupts the balance between the two losses, leading to suboptimal performance. 


\textbf{Entropic Index} is the parameter \( q \) in Tsallis entropy (Equation \ref{eq:tsallis_entropy_def}), which controls the degree of non-extensivity in the system. It adjusts how much weight is given to rare versus frequent events in the routing mechanism. When \( q = 1 \), Tsallis entropy reverts to Shannon entropy, treating all token-routing decisions uniformly. However, varying \( q \) allows the model to emphasize or de-emphasize token assignments to experts based on their likelihood, influencing the balance between exploration (specialization of experts) and exploitation (generalization). We conducted experiments by selecting the entropic index within the range of 1.0 to 1.4 and find that an entropic index of 1.1 yielded the best overall performance(Figure \ref{fig:images}(b)), suggesting that introducing Tsallis entropy rather than Shannon entropy allows for better adaptation to task complexity. Deviating from this optimal value, either by increasing or decreasing \( q \), led to reduced accuracy. This indicates the critical role of \( q \) in fine-tuning the routing strategy and balancing expert specialization with generalization.

\textbf{Entropy Threshold} defines the soft routing threshold $h_\text{threshold}$ in \model{}. Typically, tokens with higher entropy lead to unclear routing decisions. In our method, we send high-entropy tokens to all experts through soft routing, allowing them to participate in the gradient update. Therefore, setting a reasonable threshold to constrain the soft routing algorithm is crucial. We collect the performance data for models with the entropy threshold set from 0.7 to 0.95, and find that an entropy threshold of 0.9 produced the highest accuracy (Figure~\ref{fig:images}(c)). Lower thresholds led to the over-selection of experts, causing computational inefficiency without significant performance gains, while higher thresholds resulted in under-utilization of experts, limiting the model's capacity.

\subsubsection{Identifying the Optimal Hyperparameter for Effective Expert Selection}

This part include the discussion on two hyperparameters that closely related to expert selection.

\textbf{Top-p} refers to the threshold $p$ in the Top-p algorithm \citep{huang2024harder}, originally designed for MoE models. This algorithm collects the confidence level of each expert in handling input $x$ and activates a number of experts based on cumulative probability. We examined the impact of various $p$ values, ranging from 0.6 to 0.95, and found that a Top-p value of 0.75 resulted in the highest accuracy (Figure \ref{fig:images}(d)). These results highlight the importance of selecting an optimal Top-p value to balance expert specialization and generalization, while also demonstrating the superior adaptability of \model{} compared to the Top-p approach (Figure \ref{fig:enter-label}).

\textbf{Keep-Top-k} means the minimum number of activated experts $k$ in our architecture. The work on the Switch Transformer\citep{fedus2022switch} defined expert capacity, noting that if tokens are unevenly dispatched, certain experts may overflow. Due to the limited number of activated parameters, the expert capacity of MoLE is lower than that of MoE models, and during fine-tuning, activating only one expert can lead to overfitting, therefore, we tested different $k$ values ranging from 1 to 4 on the dynamic routing strategy in \model{} to explore the optimal minimum number of activated experts. We find that increasing $k$ to 2 provided the necessary parameter activation, resulting in the best performance improvements (Figure \ref{fig:images}(e)), maximally avoiding overfitting issues.

\begin{wrapfigure}{}{0.45\textwidth}
    \centering
    \includegraphics[width=\linewidth]{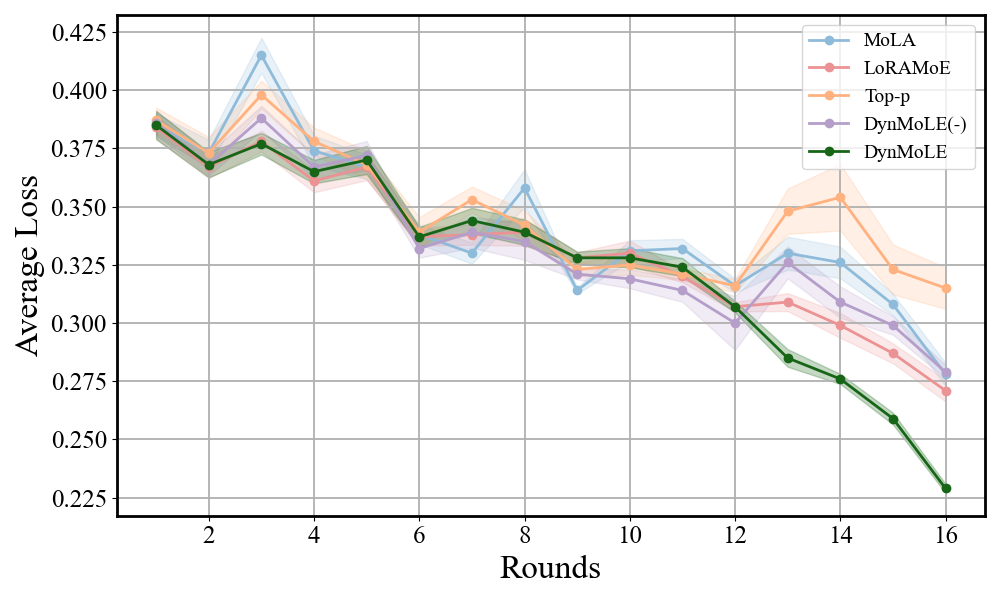}
    \caption{The entropy loss of \model{} efficiently reduces uncertainty during fine-tuning on RTE. }
    \label{fig:loss_comparison}
\end{wrapfigure}

By appropriately configuring the parameters mentioned above, the experiments demonstrated a significant improvement in \model{}'s performance. Figure~\ref{fig:loss_comparison} shows the change in loss during fine-tuning on the GLUE-RTE dataset. We treat every 320 training steps as one round and calculate the mean and standard deviation for each round, resulting in a significantly lower average loss compared to other methods 
This improvement is especially evident after 12 rounds, where \model{} outperforms MoLA, LoRAMoE, and Top-p strategies, highlighting its effectiveness in mitigating training uncertainty. Additionally, by expanding the token allocation in three dimensions, \model{} shows greater ability than the Top-p method in reducing system entropy and efficiently assigning tokens to optimal experts (refer to Appendix~\ref{appd:word_embd} for more details).

\section{Conclusion}
\label{sec:conclusion}

In this paper, we introduce \model{}, a hybrid routing strategy that enables the routing mechanism to dynamically adjust based on the entropy of the router's probability distribution for each token. This dynamic adjustment allows for more flexible and efficient expert selection, optimizing performance across diverse conditions while balancing exploration and exploitation in token routing. Our extensive experiments on commonsense reasoning benchmarks demonstrate that \model{} achieves significant performance improvements.


\clearpage


\bibliography{iclr2025_conference}
\bibliographystyle{iclr2025_conference}

\clearpage

\appendix
\section{Appendix}
\subsection{Uncertainty and Entropy}\label{appd:muomr}


Consider an MoE (Mixture of Experts) model with $N$ experts. This is an initial MoE model using only a soft routing algorithm. For an input $x$, the model's output is given by Equation~\ref{eq:vanilla_moe}.

Here, $G(x) = [G_1(x), G_2(x), \dots, G_N(x)]$ is the router's output after softmax normalization, initialized to a uniform distribution, satisfying $G_i(x) \geqslant 0$, $\sum_{i=1}^{N} G_i(x) = 1$. $E_i(x)$ represents the output of the $i$-th expert.

We define a differentiable and convex loss function $L = L(f_{\text{MoE}}(x), y)$. During training, the gating network adjusts the model parameters using gradient descent to minimize the loss function, which measures the difference between the model’s predicted output $f_{\text{MoE}}(x)$ and the true label $y$. Specifically, we update:

\begin{equation}
G_i(x) \leftarrow G_i(x) - \eta \cdot \frac{\partial L}{\partial G_i(x)}
\end{equation}

where $\eta$ is the learning rate. The partial derivative of the loss function $L$ with respect to the gating network output $G_i(x)$ is:

\begin{equation}
\frac{\partial L}{\partial G_i(x)} = \frac{\partial L}{\partial f_{\text{MoE}}(x)} \cdot \frac{\partial f_{\text{MoE}}(x)}{\partial G_i(x)} =\frac{\partial {[G_1(x)E_1(x)+\dots+G_i(x)E_i(x)+\dots+G_N(x)E_N(x)]}}{\partial G_i(x)}  = \frac{\partial L}{\partial f_{\text{MoE}}(x)} \cdot E_i(x)
\end{equation}

This shows that the gradient of the loss function $L$ with respect to $G_i(x)$ is proportional to the output of the expert $E_i(x)$:

\begin{equation}
\frac{\partial L}{\partial G_i(x)} \propto E_i(x)
\end{equation}

This indicates that the larger the influence of $E_i(x)$ on the model's output, the larger the absolute value of the gradient, meaning that adjusting $G_i(x)$ will have a more significant impact on reducing the loss. The expert’s output directly affects both the direction and magnitude of the weight update.Through gradient updates, the gating network adaptively adjusts $G_i(x)$ to increase the weights of experts that help reduce the loss and decrease the weights of experts that increase the loss.

In the ideal case, when the model fully converges and $L$ reaches its minimum, we expect for all experts:

\begin{equation}
\frac{\partial L}{\partial G_i(x)} = \frac{\partial L}{\partial f_{\text{MoE}}(x)} \cdot E_i(x) = 0 \quad \text{for all } i = 1, 2, \dots, N
\end{equation}

Since the gradient of the loss function with respect to the model output, $\frac{\partial L}{\partial f_{\text{MoE}}(x)}$, is a constant vector for a fixed input $x$. For experts that satisfy $\frac{\partial L}{\partial f_{\text{MoE}}(x)} \cdot E_i(x) = 0$, there may be non-zero $G_i(x)$; for other experts where $\frac{\partial L}{\partial f_{\text{MoE}}(x)} \cdot E_i(x) \neq 0$, i.e., experts that cannot minimize the loss, $G_i(x)$ must converge to zero to satisfy the zero-gradient condition.

This analysis shows that the distribution of $G(x)$ will tend to assign a value of 1 to the optimal set of experts and 0 to the other experts, forming an indicative distribution with characteristic function:

\begin{equation}
G_i(x) = 
\begin{cases} 
1, & i \in \underset{j}{\arg\min} \, L(E_j(x), y) \\
0, & \text{otherwise} 
\end{cases}
\end{equation}

However, in practice, due to factors such as regularization and numerical stability, $G(x)$ cannot fully form a perfectly indicative distribution. Moreover, an overly indicative distribution may lead to overfitting. Nonetheless, the overall trend is still to assign greater weights to a few more optimal experts, resulting in a peak-like distribution. As the probability distribution increasingly approaches this ideal peak distribution, the MoE model is often able to select the optimal set of experts. Since it is difficult to define the ideal peak distribution in an analytical mathematical form, traditional methods such as KL divergence cannot accurately measure this deviation.
\subsection{Tsallis Entropy vs. Shannon Entropy}

\subsubsection{More Flexible}\label{appd:flex}

\begin{align}
S_q(p) = \frac{1 - \sum_{i=1}^N p_i^q}{q - 1}
\end{align}
we employ a Taylor series expansion around $ q = 1 $. Consider the function $ f(q) = p_i^q $ and expand it around $ q = 1 $:
\begin{align}
p_i^q &= p_i^{1 + (q - 1)} \nonumber \\
&= p_i \cdot p_i^{q - 1} \nonumber \\
&= p_i \exp\left[ (q - 1) \log p_i \right] \nonumber \\
&= p_i \left[ 1 + (q - 1) \log p_i + \frac{1}{2} (q - 1)^2 (\log p_i)^2 + \cdots \right].
\end{align}

Sum over all $ i $:
\begin{align}
\sum_{i=1}^N p_i^q &= \sum_{i=1}^N \left[ p_i + p_i (q - 1) \log p_i + \frac{1}{2} p_i (q - 1)^2 (\log p_i)^2 + \cdots \right] \nonumber \\
&= 1 + (q - 1) \sum_{i=1}^N p_i \log p_i + \frac{1}{2} (q - 1)^2 \sum_{i=1}^N p_i (\log p_i)^2 + \cdots.
\end{align}

Using the normalization condition $ \sum_{i=1}^N p_i = 1 $, we have:
\begin{align}
1 - \sum_{i=1}^N p_i^q &= - (q - 1) \sum_{i=1}^N p_i \log p_i - \frac{1}{2} (q - 1)^2 \sum_{i=1}^N p_i (\log p_i)^2 + \cdots.
\end{align}
Substituting back into the definition of the Tsallis entropy:
\begin{align}
S_q(p) &= \frac{1 - \sum_{i=1}^N p_i^q}{q - 1} \nonumber \\
&= -\sum_{i=1}^N p_i \log p_i - \frac{1}{2} (q - 1) \sum_{i=1}^N p_i (\log p_i)^2 + \cdots.
\end{align}

As $ q \to 1^+ $, the term $ (q - 1) $ approaches zero, and higher-order terms become negligible. Thus, we obtain:
\begin{align}
\lim_{q \to 1^+} S_q(p) = -\sum_{i=1}^N p_i \log p_i.
\end{align}

\subsubsection{More Stable}

Consider a loss function that incorporates an entropy regularization term:
\begin{equation}
    \mathcal{L} = \mathcal{L}_{\text{data}} - \lambda \cdot \text{Entropy}(f_{\text{MoE}}(x)),
\end{equation}
where $\lambda$ is a regularization coefficient controlling the influence of entropy on the overall loss, and $\mathcal{L}_{\text{data}}$ represents the data loss, which measures the discrepancy between model predictions and the true labels. $\text{Entropy}(f_{\text{MoE}}(x))$ quantifies the uncertainty in the router's output.

Now, let us compare the loss functions using Shannon entropy and Tsallis entropy, respectively:

\begin{align}
    \mathcal{L}_{\text{Shannon}} =& \mathcal{L}_{\text{data}} - \lambda \cdot \sum_{i=1}^n G_i(x) \log G_i(x), \\
    \mathcal{L}_{\text{Tsallis}} =& \mathcal{L}_{\text{data}} - \lambda \cdot \frac{1}{q - 1} \left( 1 - \sum_{i=1}^n G_i(x)^q \right),
\end{align}
where $G_i(x)$ represents the routing probability of expert $i$.

When optimizing these loss functions via gradient descent, the gradients with respect to $G_i(x)$ are given by:

\begin{align}
    \frac{\partial \mathcal{L}_{\text{Shannon}}}{\partial G_i} =& \frac{\partial \mathcal{L}_{\text{data}}}{\partial G_i} - \lambda \cdot\left(1 + \log G_i\right), \\
    \frac{\partial \mathcal{L}_{\text{Tsallis}}}{\partial G_i} =& \frac{\partial \mathcal{L}_{\text{data}}}{\partial G_i} - \lambda \cdot G_i^{q - 1}.
\end{align}

For Shannon entropy, as $G_i(x) \to 0$, $\log G_i \to -\infty$, which can lead to steep gradient magnitudes and unstable updates. In contrast, for Tsallis entropy, as $G_i(x) \to 0$, the gradient $\lambda G_i^{q - 1} \to 0$ ($q > 1$), reducing the impact of low-probability events and providing a more stable optimization process.

\subsection{Datasets} \label{appd:dataset}
Table~\ref{tab:dataset_description} presents detailed information about the datasets used in our experiments, including their task names, respective domains, the number of training and test sets, task types. All datasets are downloaded from \href{https://huggingface.co}{HuggingFace} by using the \textsc{Datasets} library in Python.
\begin{table}[!h]
\centering
\caption{Description of Datasets used in experiments.}
\begin{tabular}{l|lccr}
\toprule
\textbf{Task Name} & \textbf{Domain} & \textbf{\# Train} & \textbf{\# Test} & \textbf{Task Type}\\
\midrule
RTE & GLUE Benchmark & 2,490 & 277 & Textual Entailment \\
BoolQ & Wikipedia & 9,427 & 3,270 & Text Classification \\
ARC-E & Natural Science & 2,250 & 2,380 & Question Answering \\
ARC-C & Natural Science & 1,120 & 1,170 & Question Answering \\
OpenBookQA & Science Facts & 4,957 & 500 & Question Answering \\
PIQA & Physical Interaction & 16,100 & 1,840 & Question Answering \\
SIQA & Social Interaction & 33,410 & 1,954 & Question Answering \\
HellaSwag & Video Caption & 39,905 & 10,042 & Sentence Completion \\
WinoGrande & Winograd Schemas & 9,248 & 1,267 & Fill in the Blank \\
\bottomrule
\end{tabular}
\label{tab:dataset_description}
\end{table}

\subsection{Hyper Parameters Setting} \label{appd:hyper-p}
\begin{table}[!h]
\caption{Hyperparameter configurations for all baseline methods and \model{} fine-tuning with LLaMA2-7B.}
\label{table:hyperparameters}
\centering
\begin{tabular}{l|cccc}
\toprule
\textbf{Hyperparameters}      & \textbf{LoRA/DoRA} & \textbf{LoRAMoE} & \textbf{MoLA} & \textbf{\model}   \\
\midrule
Cutoff Length & \multicolumn{4}{c}{512} \\
Learning Rate & \multicolumn{4}{c}{2e-4} \\
Optimizer     & \multicolumn{4}{c}{AdamW} \\
Batch size    & \multicolumn{4}{c}{16} \\
Accumulation Steps & \multicolumn{4}{c}{8} \\
Dropout       & \multicolumn{4}{c}{0.05} \\
Epochs        & \multicolumn{4}{c}{2} \\
Where         & \multicolumn{4}{c}{Up, Down, Gate} \\
\midrule
LoRA Rank $r$           & 80 & 24 & 24 & 24 \\
LoRA Alpha $\alpha$     & 160 & 48 & 48 & 48 \\
Experts       & - & 6 & 6 & 6 \\
Top-K         & - & - & 2 & 2 \\
Top-P         & - & - & - & 0.75 \\
Entropy Threshold       & - & - & - & 0.9 \\
Entropy Index           & - & - & - & 1.1 \\
\bottomrule
\end{tabular}
\label{table:hyper_params}
\end{table}

\subsection{Flexibility Studies}
\label{appd:flex_studies}


\begin{figure}[!h]
    \centering
    \resizebox{0.8\textwidth}{!}{
        \begin{minipage}{0.475\linewidth}
            \includegraphics[width=\linewidth]{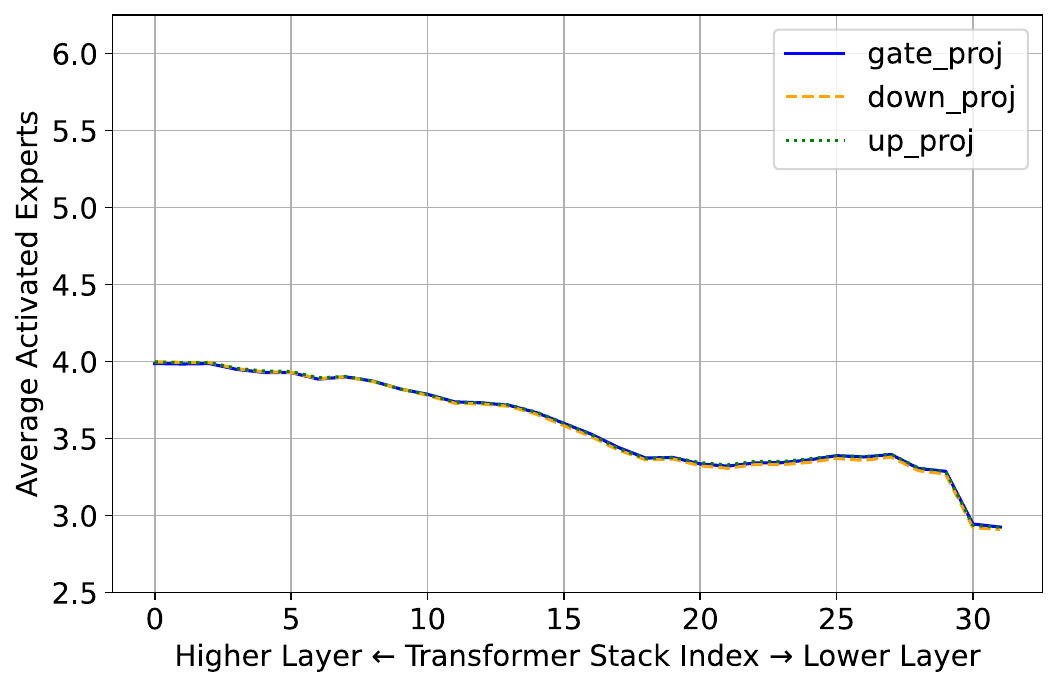}
            \subcaption{Top-P Routing Strategy}
        \end{minipage}%
        \begin{minipage}{0.475\linewidth}
            \includegraphics[width=\linewidth]{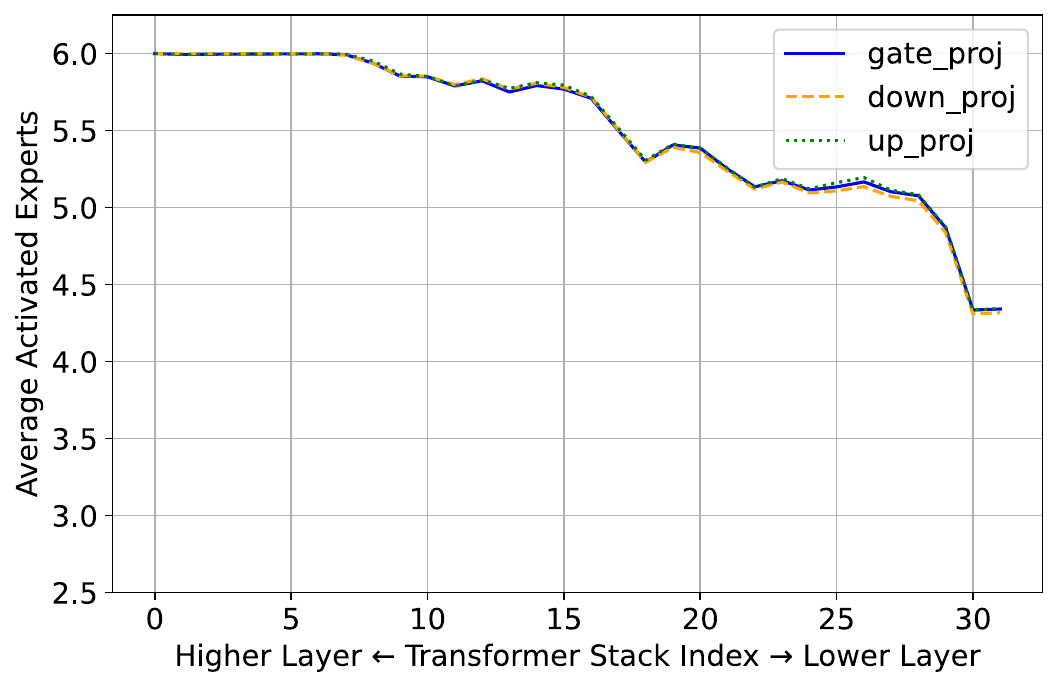}
            \subcaption{\model{}}
        \end{minipage}%
    }
    \caption{The average number of activated experts across different transformer layers.}
    \label{fig:enter-label}
\end{figure}

We evaluated the flexibility of \model{} and the Top-p method on the ARC-c dataset (as shown in Figure~\ref{fig:enter-label}). Similar to Top-p, \model{} demonstrates comparable performance across the three projections. However, \model{} activates the appropriate number of experts earlier and more comprehensively in response to router uncertainty.

\subsection{Word Embedding}\label{appd:word_embd}
\begin{figure}[h]
    \centering
    \begin{minipage}{0.3\textwidth}
        \centering
        \includegraphics[width=\linewidth]{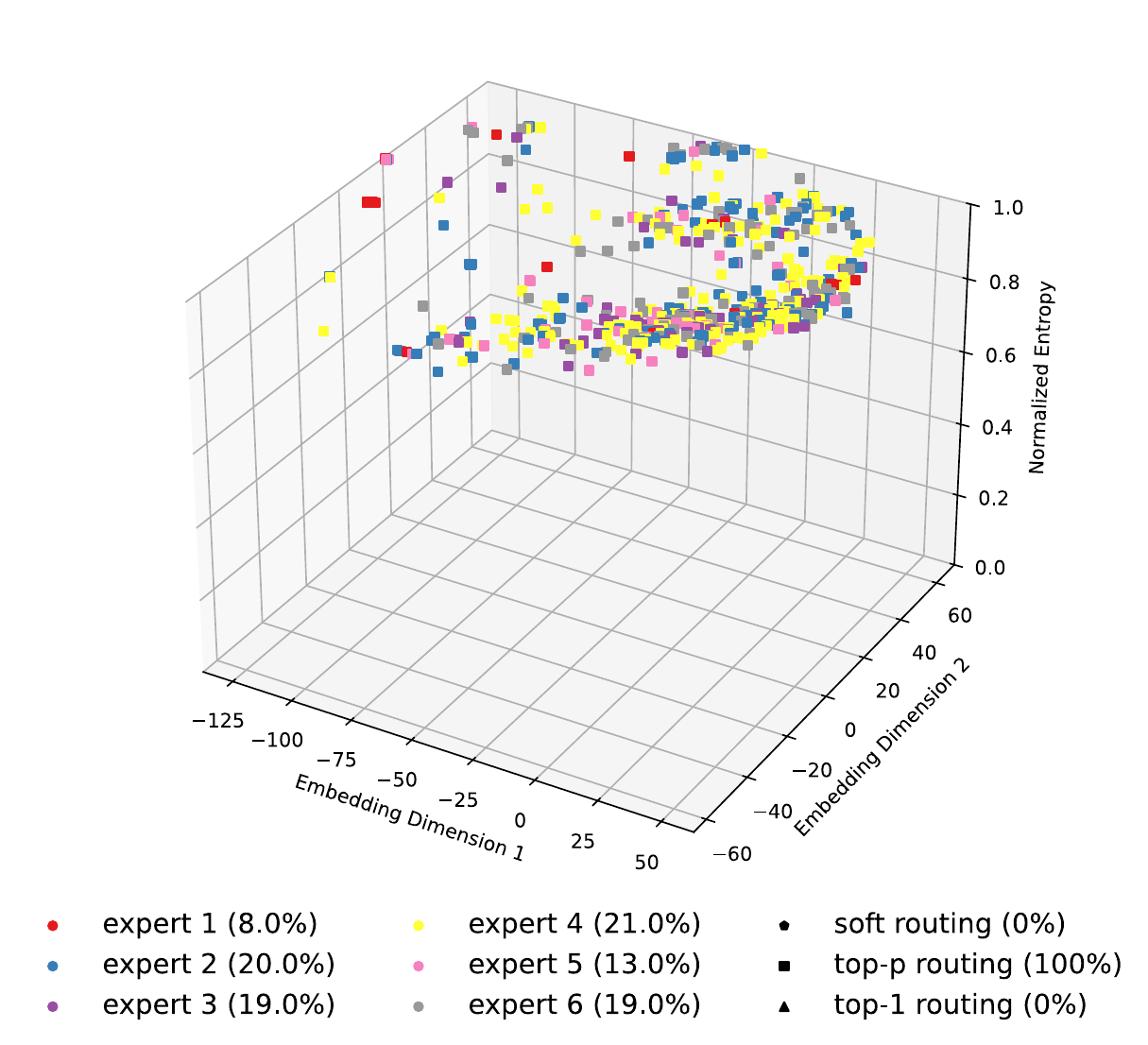}
        \subcaption{Top-P Routing of Layer 1}
    \end{minipage}
    \begin{minipage}{0.3\textwidth}
        \centering
        \includegraphics[width=\linewidth]{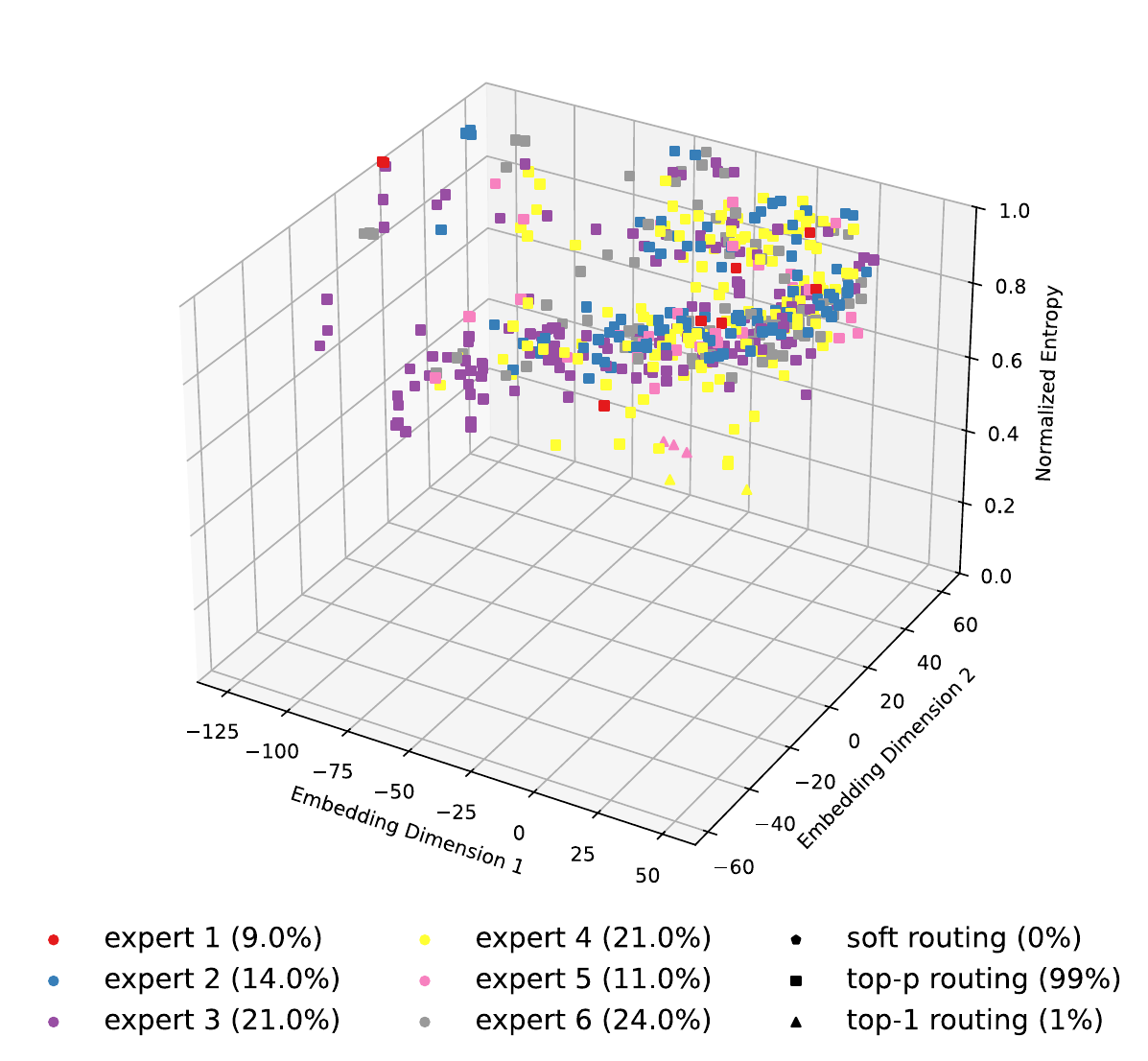}
        \subcaption{Top-P Routing of Layer 16}
    \end{minipage}
    \begin{minipage}{0.3\textwidth}
        \centering
        \includegraphics[width=\linewidth]{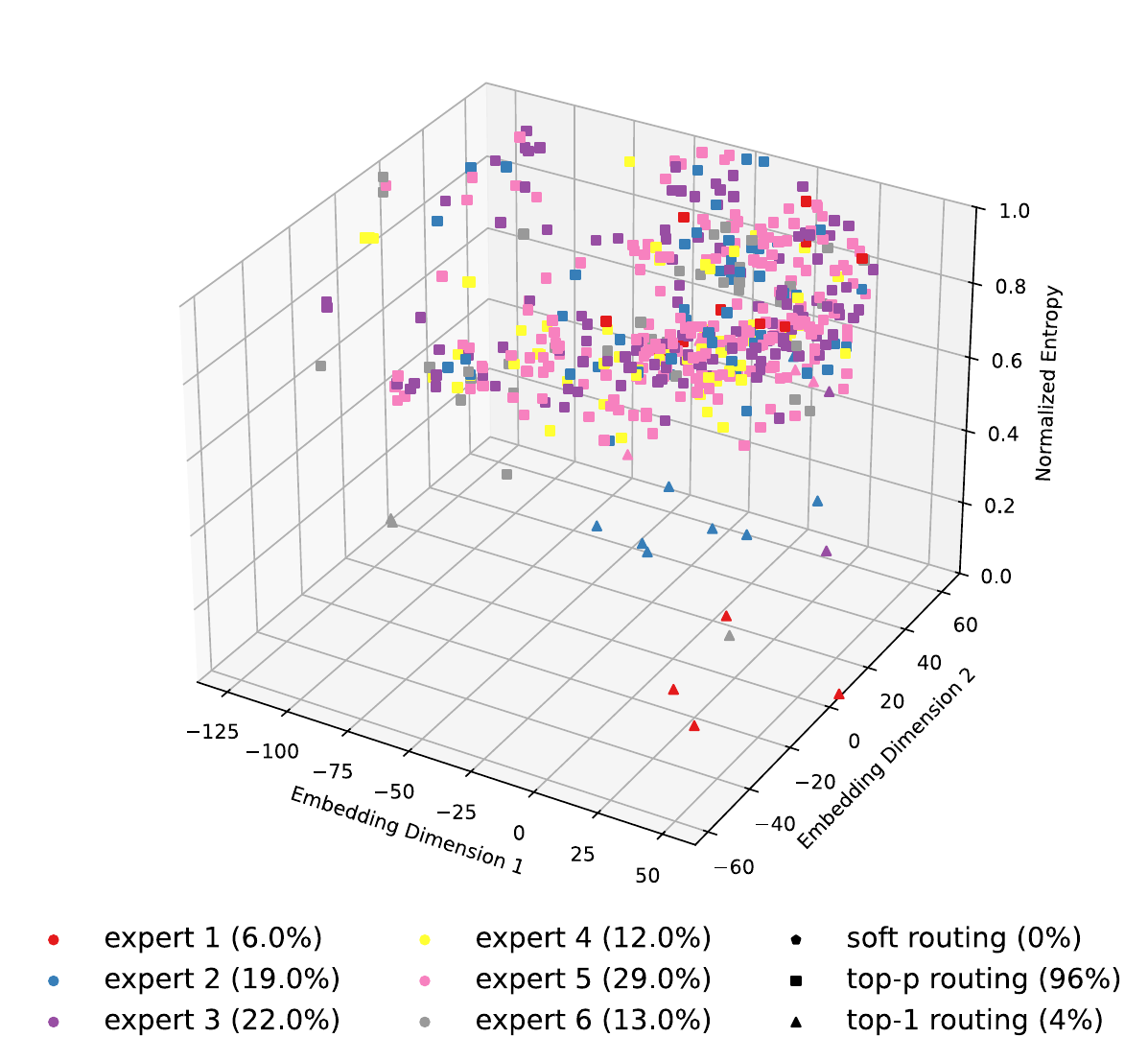}
        \subcaption{Top-P Routing of Layer 32}
    \end{minipage}
    \begin{minipage}{0.3\textwidth}
        \centering
        \includegraphics[width=\linewidth]{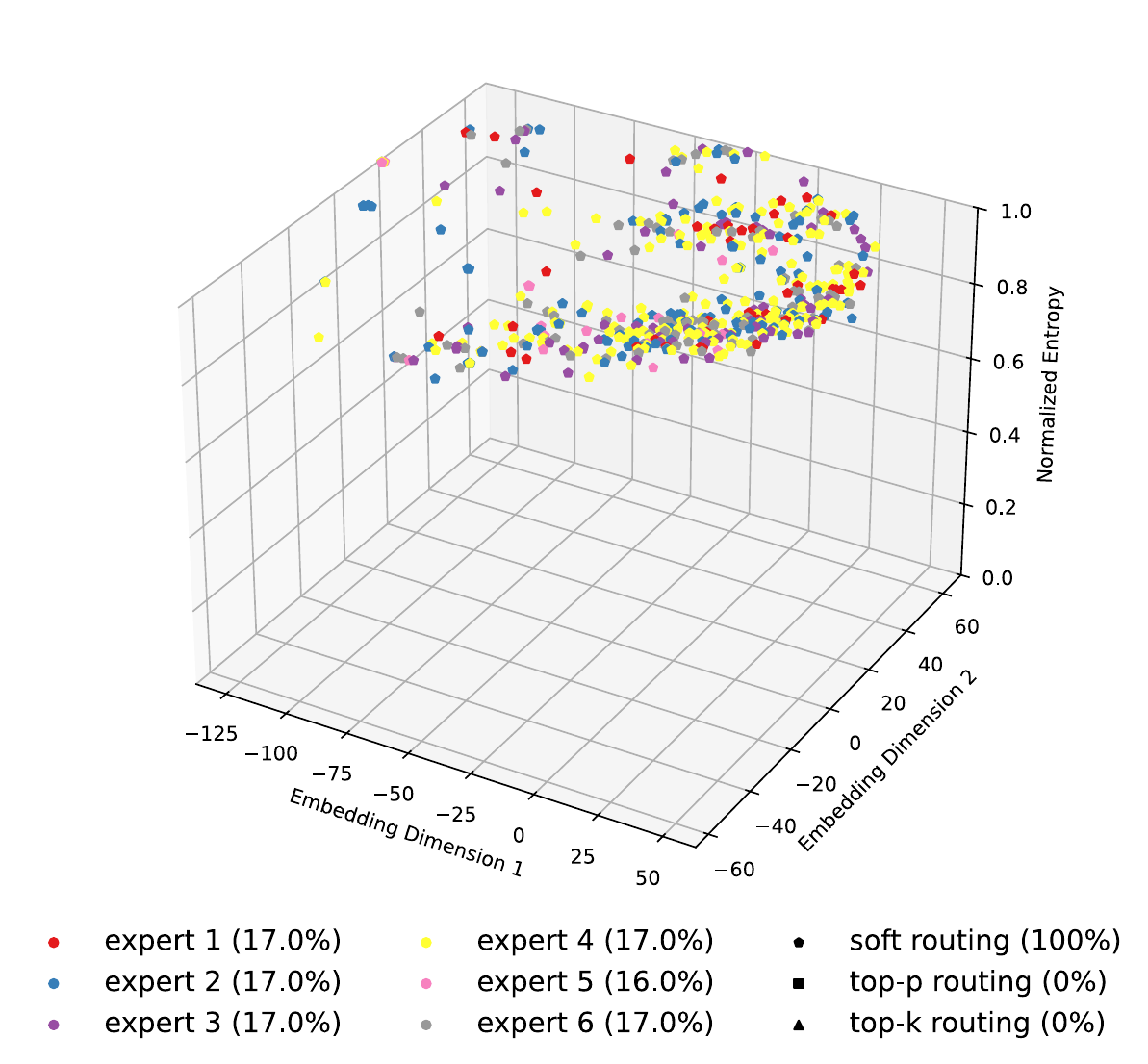}
        \subcaption{\model{} of Layer 1}
    \end{minipage}
    \begin{minipage}{0.3\textwidth}
        \centering
        \includegraphics[width=\linewidth]{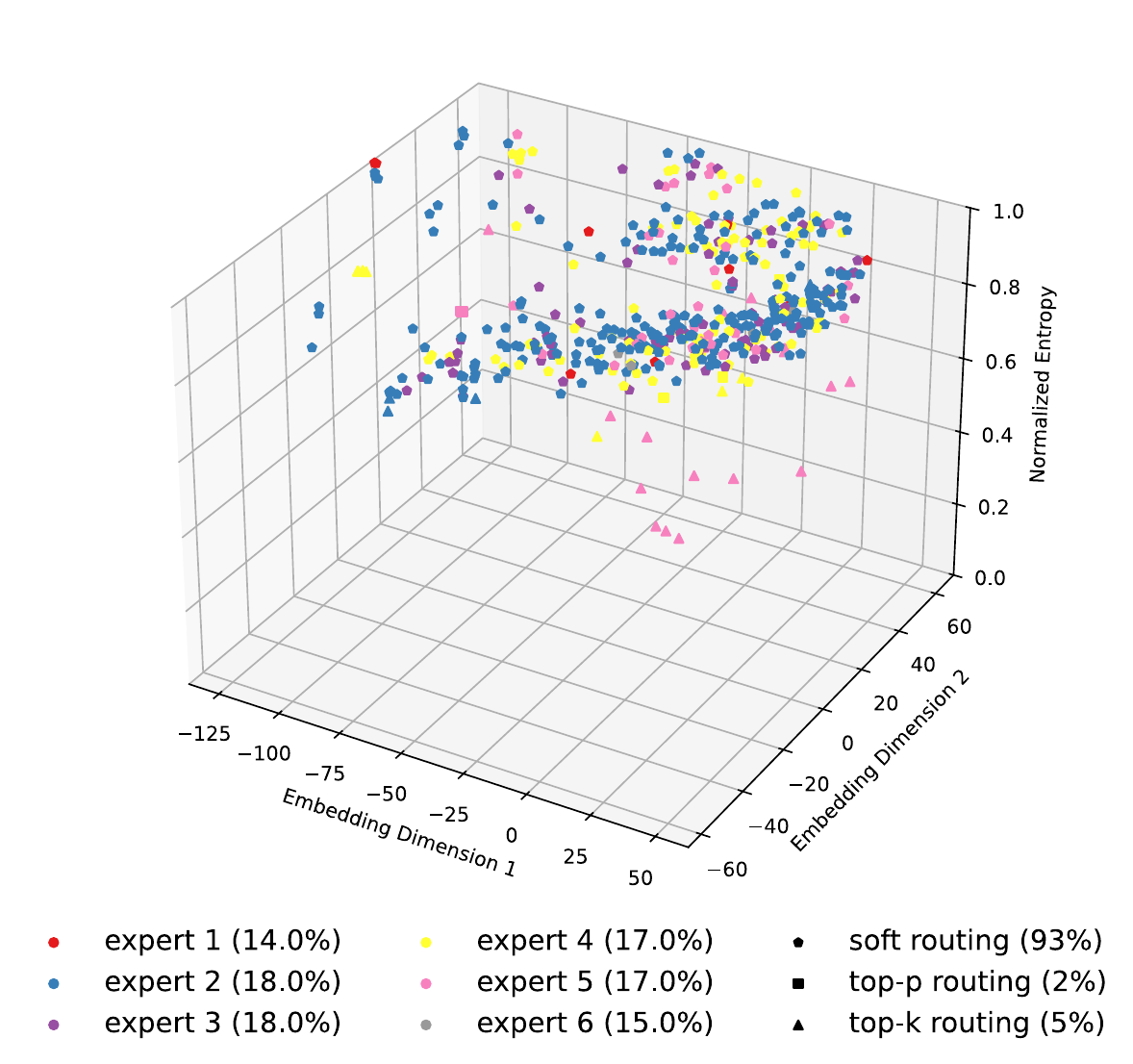}
        \subcaption{\model{} of Layer 16}
    \end{minipage}
    \begin{minipage}{0.3\textwidth}
        \centering
        \includegraphics[width=\linewidth]{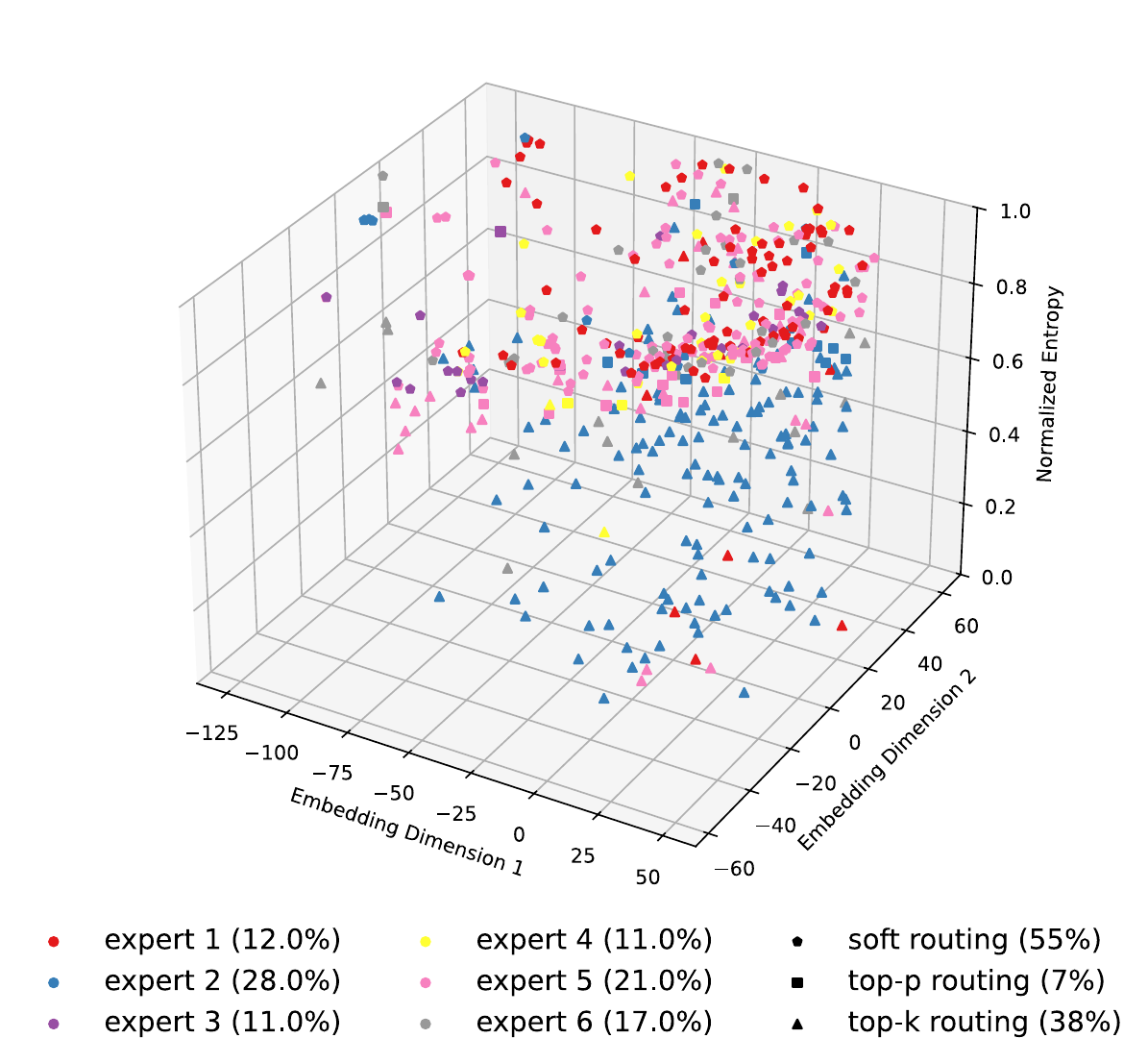}
        \subcaption{\model{} of Layer 32}
    \end{minipage}
    \caption{3D visualization of token embeddings and router entropy for each token.}
    \label{fig:layer_3d}
\end{figure}

As shown in Figure~\ref{fig:layer_3d}, we present a visualization comparing token allocation between the Top-P routing strategy~\citep{huang2024harder} and our proposed \model{} approach. We embedded 2D visualized word tokens from 10 randomly selected sentences, coloring them based on their most confidently routed expert index from the 1st, 16th, and 32nd Transformer layers, and projected them into 3D space using their normalized entropy. The percentage distribution of routing strategies is shown for each method. Compared to Top-P, \model{} more efficiently routes tokens with similar entropy to similar experts, resulting in significantly lower average entropy and a more balanced load across experts, as reflected in the more even distribution across layers.

\end{document}